\documentclass[runningheads]{llncs}
\usepackage{booktabs}
\usepackage{multirow}
\usepackage{graphicx}
\usepackage{marvosym}

\usepackage{graphicx} 
\begin{document}
\title{MMF: Multi-Task Multi-Structure Fusion for Hierarchical Image Classification\thanks{ Supported by the National Natural Science Foundation of China (No. 62006221), the Open Research Project of the State Key Laboratory of Media Convergence and Communication, Communication University of China, China (No. SKLMCC2020KF004), the Beijing Municipal Science \& Technology Commission (Z191100007119002), and the Key Research Program of Frontier Sciences, CAS, Grant NO ZDBS-LY-7024.}}
%
%
\author{Xiaoni Li\inst{1,2} \and
Yucan Zhou\inst{1,}\textsuperscript{\Letter} \and
Yu Zhou\inst{1} \and 
Weiping Wang\inst{1}}
\authorrunning{X. Li et al.}
\titlerunning{MMF for Hierarchical Image Classification}
\institute{Institute of Information Engineering, Chinese Academy of Sciences, Beijing, China \and
School of Cyber Security, University of Chinese Academy of Sciences, Beijing, China \\
\email{\{lixiaoni, zhouyucan, zhouyu, wangweiping\}@iie.ac.cn}}
\maketitle              
\begin{abstract}
Hierarchical classification is significant for complex tasks by providing multi-granular predictions and encouraging better mistakes.  As the label structure decides its performance, many existing approaches attempt to construct an excellent label structure for promoting the classification results. In this paper, we consider that different label structures provide a variety of prior knowledge for category recognition, thus fusing them is helpful to achieve better hierarchical classification results. Furthermore, we propose a multi-task multi-structure fusion model to integrate different label structures. It contains two kinds of branches: one is the traditional classification branch to classify the common subclasses, the other is responsible for identifying the heterogeneous superclasses defined by different label structures. Besides the effect of multiple label structures, we also explore the architecture of the deep model for better hierachical classification and adjust the hierarchical evaluation metrics for multiple label structures. Experimental results on CIFAR100 and Car196 show that our method obtains significantly better results than using a flat classifier or a hierarchical classifier with any single label structure.

\end{abstract}

\section{Introduction}

\begin{figure*}[htb] 
 \center{\includegraphics[width=0.9\textwidth] {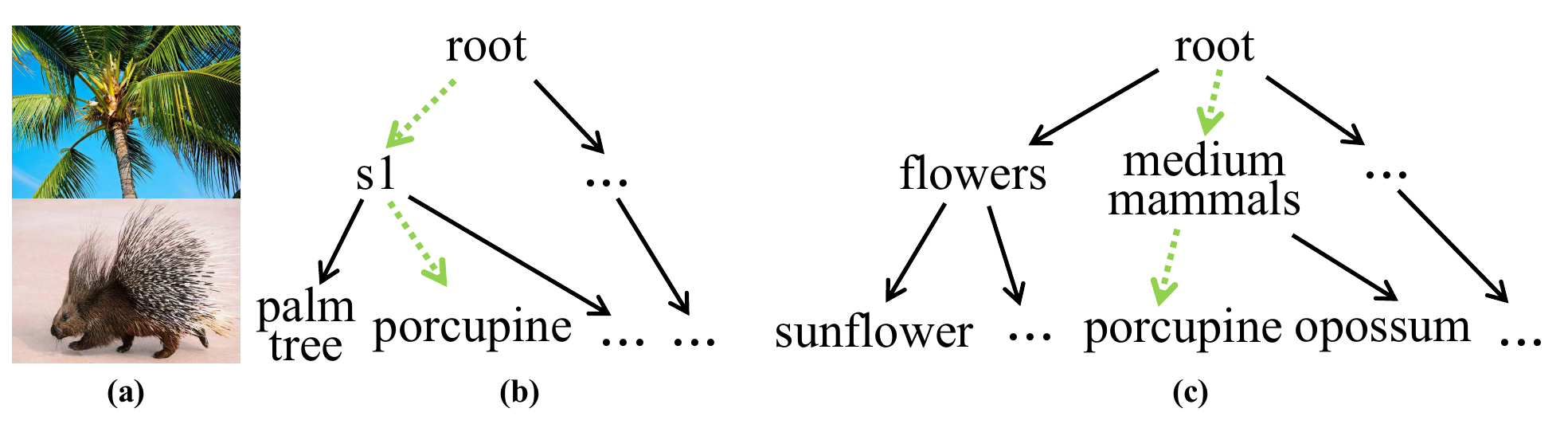}}
 \caption{\label{1} The benefit of combining multiple label structures. (a) shows two samples for palm tree and porcupine, as these two categories are similar, they are easy to be misclassified. (b) and (c) are two label structures, each for the visual sturcture based on the affinity matrix and the semantic structure. The second layers in (b) and (c) are the superclasses, the third layers are the shared subclasses. The green dashed paths are the ground truth in both label structures when a ``porcupine” is needed to identify. When ``s1”and ``medium mammals”are recognized in each label structure, ``porcupine”is promoted as it belongs to both ``s1”and ``medium mammals”. } 
 \end{figure*}

Although deep learning in text spotting \cite{qxg1,qxg2,cyd1,cyd2,qz1,qz2}, object detection \cite{ydb1},  self-supervised learning \cite{zyf1,zyf2,ldz1,ldz2,ldz3} and image classification \cite{DBLP:journals/corr/abs-1708-04552,DBLP:conf/iccvw/Krause0DF13} has achieved dramatic performance with the increase of annotated data, the unclassifiable categories are growing and inevitable in the ear of big data. Moreover, the conventional one-hot coding in flat classifiers suggests a strict error evaluation: as long as the predicted value is inconsistent with the real one, it will be recognized as misclassification. In fact, there are different levels of severity in mistakes \cite{DBLP:journals/corr/abs-1912-09393}. As shown in Figure 1(c), the classifier makes a less serious mistake when it classifies a ``porcupine" into an ``opossum" than a ``
sunflower" obviously, because they all belong to the superclass ``medium mammals". Therefore, when misclassification is unavoidable, providing a reasonable mistake is more significant.

Recently, more and more work is devoted to using hierarchical classification methods \cite{wang2021hierarchical,8809246} to make multi-granular predictions and avoid serious mistakes. In hierarchical classification, label structures play a critical role. Hence many researchers try to construct efficient label structures, which can be roughly divided into semantics-based methods and computation-based methods.  The former extracts the semantic structure from WordNet \cite{Lin1999WordNet}, where categories are organized into a tree-shape structure according to their semantic relations \cite{DBLP:conf/nips/ZhaoLX11,DBLP:conf/cvpr/DengDSLL009,DBLP:conf/cvpr/DengKBF12,DBLP:conf/cvpr/GuillauminF12}. However, these relations may be inconsistent with the appearances, which weakens the performance of classification tasks.  Therefore, a lot of work builds visual information tree structures \cite{DBLP:conf/cvpr/GriffinP08,DBLP:conf/nips/BengioWG10,DBLP:conf/cvpr/LiuSTSL13,DBLP:journals/corr/abs-1906-02012,DBLP:journals/corr/abs-1906-01536,DBLP:journals/pr/LeiMZDZF14,DBLP:journals/tip/FanZPG15,DBLP:journals/tip/QuLSLWXT17}. Some build the tree structures based on the confusion matrix \cite{DBLP:conf/cvpr/GriffinP08,DBLP:conf/nips/BengioWG10,DBLP:conf/cvpr/LiuSTSL13,DBLP:journals/corr/abs-1906-02012,DBLP:journals/corr/abs-1906-01536}, which is constructed by the results of a classifier. Others construct the label structure based on the affinity matrix \cite{DBLP:journals/pr/LeiMZDZF14,DBLP:journals/tip/FanZPG15,DBLP:journals/tip/QuLSLWXT17} calculted by the similarity of any two categories.

Different label structures provide various prior knowledge for the underlying classification tasks. Hence integrating these structures can further improve the performance \cite{DBLP:conf/icmcs/WangF13,zhao2016fusing}. As shown in Fig.1, in the mission of ``porcupine" classification, if one has determined its superclass ``s1" and ``medium mammals" according to the label structure based on the affinity matrix and semantics respectively, then, ``porcupine" can be easily determined by combining these two intermediate results. A straightforward strategy to fuse multiple label structures is constructing a hierarchical classifier for each structure, and the prediction is obtained by integrating the results of multiple classifiers \cite{DBLP:conf/icmcs/WangF13}. This idea is simple and efficient, but in the deep learning scenario, it is memory-consuming and computationally redundant to design a neural network for each label structure. 

In this paper, a multi-task multi-structure fusion (MMF) model is proposed to make the superclasses from different label structures instruct the subclass recognition. It achieves this by encouraging the learned feature to satisfy the multiple similarity constraints in various hierarchical label structures. Specifically, it is a deep convolutional neural network with two kinds of classification branches: the conventional classification branch (CCB) used for identifying subclasses, and the multiple superclass classification branches (MSCBs), where each branch is responsible for recognizing the superclasses defined by a specific label structure. 

Our main contributions are summarized in three folds: 1) We find that integrating multiple label structures can further improve the performance of hierarchical classification, and propose a MMF model to combine different hierarchical label structures. 2) Further, various architectures of our MMF model are explored for better classification. 3) We adjust the hierarchical evaluation metrics for multiple label structures. Experimental results on CIFAR100 and Car196 are better than traditional flat classifiers and hierarchical classifiers with any single label structure.

\vspace{-2px}
\section{ Related Work}
\vspace{-3px}

\subsection{Hierarchical Classification}
The traditional methods decompose the hierarchical classification task into several subtasks and train a subclass classifier for each superclass node independently \cite{DBLP:conf/cvpr/GriffinP08,DBLP:conf/nips/BengioWG10,DBLP:journals/tip/QuLSLWXT17,DBLP:conf/nips/DengSBL11}. However, this strategy is memory-consuming and computing expensive for storing and training many subclass classifiers. Therefore, these methods are not suitable for deep learning. For deep hierarchical classification, Frome \textit{et al.} \cite{DBLP:conf/nips/FromeCSBDRM13} constructs a deep visual-semantic model by re-training the lower layers of the pre-trained visual network to predict the vector representation of the image label text in the hierarchical label structure learned by the language model. Barz \& Denzler \cite{DBLP:conf/wacv/BarzD19} design an algorithm to map the labels into a unit hypersphere where the cosine distances between different labels are equal to the distance in the hierarchical label structure.

\begin{figure*}[htb] 
 \center{\includegraphics[width=\textwidth] {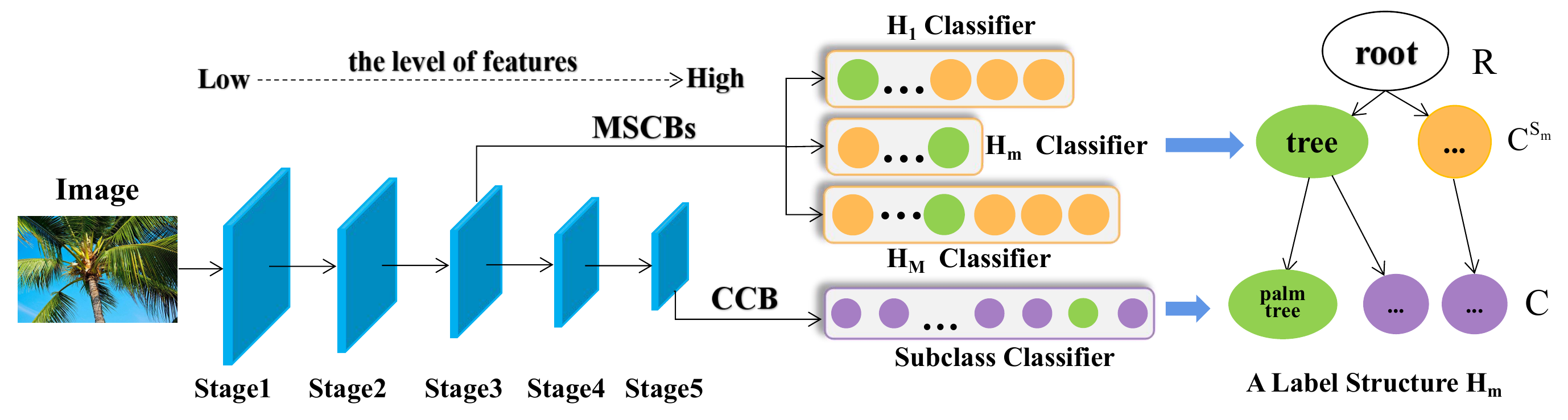}}
 \caption{\label{1}  The architecture of our MMF model with a five-stage CNN.} 
 \end{figure*}
 
 Besides the implicit label embedding, many researchers want to explicitly model the hierarchical label structure. Wu \textit{et al.} \cite{DBLP:conf/mm/WuMUS16} adds one fully connected softmax layer for each layer in the hierarchy to make the network recognize both the superclasses and the subclasses. But in this work, the relations between the superclasses and subclasses are underutilized. Bertinetto \textit{et al.} \cite{DBLP:journals/corr/abs-1912-09393} adds a weight matrix between the superclass classifier and the subclass classifier, thus, predictions of the superclass can be propagated to and affect the predictions of the subclass through the weight matrix.  Ahmed \textit{et al.} \cite{DBLP:conf/eccv/AhmedBT16} trains a network to provide superclasses information and common knowledge through shared features to a set of expert networks, each of which devoted to recognizing the subclasses of a specific superclass. Therefore, the multi-task framework has been proved efficient for hierarchical classification.

  \vspace{-10px}
\subsection{Multiple Label Structures Fusion}

 As we have mentioned, different label structures provide different priori knowledge for hierarchical classification, thus, integrating these structures can further improve the performance. Wang \textit{et al.} \cite{DBLP:conf/icmcs/WangF13} constructs a classifier for each label structure, and the prediction of a test sample is obtained by integrating the results of multiple classifiers. This idea is simple and efficient, but in the deep learning scenario, it is memory-consuming and computationally redundant to design a neural network for each label structure. Instead of training multiple subclass classifiers, Zhao \textit{et al.} \cite{zhao2016fusing} fuses multiple category similarities defined by different label structures in the kernel space, then trains one kernel SVM classifier. Inspired by this idea, we propose a multi-task multi-structure framework to make the superclasses from different label structures instruct the subclass recognition by encouraging the learned features to satisfy the multiple similarity constraints in different label structures.
 
 \vspace{-3px}
 
 \section{Method}
 \vspace{-5px}
  \subsection{Problem Definition}
   \vspace{-3px}
Given an image dataset $\mathbf{D}$ with $N$ classes, after $M$ label structure construction methods applied, we can obtain $M$ tree-like label structures. Except for the layer containing the root node, each layer in the structure is equipped with a specific classifier to decide the category in the current layer. To simplify the problem, all the structures covered in this paper are arranged with three levels. Take the right side of Fig.2 as an example, a label structure is represented as $\mathbf{H_{m}} = \{\mathbf{R},\mathbf{C_{}^{S_{m}}},\mathbf{C}\}$, where $\mathbf{R}$ is the root node, $\mathbf{C_{}^{S_{m}}}$ is the superclass set in $\mathbf{H_{m}}$, and $\mathbf{C}$ is the subclass set. Consequently, given a sample $x$ in $\mathbf{D}$, its labels compose of one subclass $c$ and $M$ superclasses $c_{}^{s_{m}}$. For hierarchical classification with multiple label structures, all these superclasses should be predicted.

  \label{sec:intro}
 \begin{figure*}[htb] 
 \center{\includegraphics[width=0.9\textwidth] {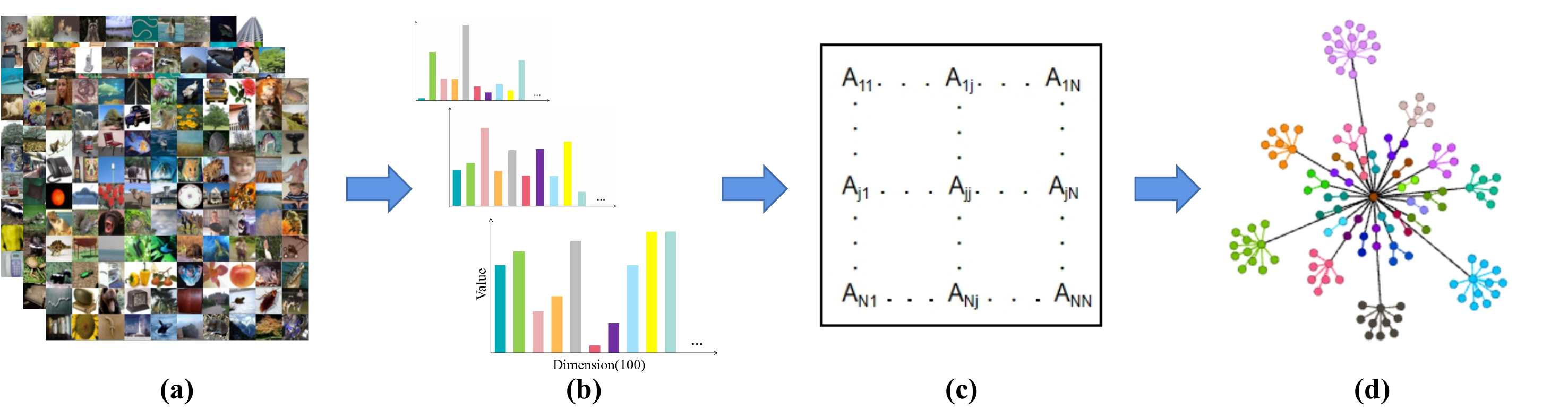}}
 \caption{\label{2} The construction of $\rm \mathbf{H_{A}}$ on CIFAR100. (a) Samples from CIFAR100 dataset. (b) Feature representations devided from 100 classes in images. (c) The affinity matrix obtained from the features. (d) $\rm \mathbf{H_{A}}$ with 30 superclasses after spectral clustering.} 
 \end{figure*}
\vspace{-2px}

\subsection{Multi-Task Multi-Structure Fusion Framework}
\vspace{-3px}
Assumed that multiple label structures have been obtained, the MMF model can be constructed, which includes two kinds of classification branches: CCB is a  classification branch with a traditional classifier to identify the subclass, while MSCBs are several classifiers for the superclass identification. Fig.2 shows an overview of our model. All MSCBs work in parallel to encourage the features derived from the network to meet various similarity constraints in different label structures, guiding CCB to make more accurate predictions.

{\bf MSCBs}
The MSCBs contains multiple superclass classifiers. As shown in Fig.2, the ``$\mathbf{H_{m}}$ classifier" in MSCBs completes its task based on the label structure $\mathbf{H_{m}}$. As the superclasses are more generic than the subclasses (e.g., medium mammals and porcupine in Fig.1 (c)), and the high-level features usually contain more details to discriminate the subclasses, MSCBs should be inserted in the early stages. We will explore the influence of various network stages to attach the MSCBs in the following experiments.

\subsection{Multiple Label Structures}
 For a dataset $\mathbf{D}$, we introduce two kinds of $\rm \mathbf{H_{m}}$ to construct our MMF model: the semantic label structure $\rm \mathbf{H_{S}}$ and the visual label structure $\rm \mathbf{H_{A}}$ based on the affinity matrix.
 
{\bf $\rm \mathbf{H_{S}}$}
Usually, a semantic structure is adopted to organize the data. Take CIFAR100 as an example: there is a three-level semantic hierarchical label structure (one root node, 20 superclasses, and 100 subclasses) in the dataset (like $\rm \mathbf{H_{m}}$ in Fig.2). We adopt this semantic structure $\rm \mathbf{H_{S}}$ inherent in the datasets in our paper.

{\bf $\rm \mathbf{H_{A}}$}
We construct a visual label structure based on the affinity matrix through two stages as shown in Fig.3: feature extraction and label structure construction. For feature extraction (from (a) to (b)), we use a pre-trained VGG16 to extract features. Then, for label structure construction, we adopt sample pairwise distance to calculate the similarity between any two categories (from (b) to (c)), and simplify the calculation with \cite{DBLP:journals/tip/QuLSLWXT17} by Eq.(1), where $c_{i}$ is the i-th class, and $Q_{c_{i}}$, $\sigma _{c_{i}}$ are the mean and variance of features in $c_{i}$. Then the affinity matrix $\rm \mathbf{A}$ can be constructed by Eq.(2), where $\delta _{ij}$ is a self-tuning parameter \cite{DBLP:journals/mta/QuWLXW14}, and we take 1 in our work. Finally, we use spectral clustering \cite{DBLP:conf/nips/NgJW01} to build the corresponding $\rm \mathbf{H_{A}}$ (from (c) to (d)).  
\vspace{-3px}
 \begin{equation}
 dis(c_{i},c_{j})_{}^{2}=\left \| Q_{c_{i}}-Q_{c_{j}} \right \|^2+\sigma _{c_{i}}^2+\sigma _{c_{j}}^2.
 \end{equation}
 
 \begin{equation}
 A_{ij}=\exp(-\frac{dis(c_{i},c_{j})}{\delta _{ij}}).
 \end{equation}
 \vspace{-14px}

\subsection{Hierarchical Measures}
As it is a multi-structure fusion work, we add the hierarchical information to the evaluation measures, and consider the similarity between the predicted class and the ground truth, \textit{i.e.}, the severity of the classifier's mistakes. However, the existing evaluation measures \cite{Zhao2020Recursive} such as hierarchical $\rm F_{1}$-measure ($F_{H}$), the tree induced loss ($TIE$) and the lowest common ancestor ($LCA$) of the prediction and the ground truth, are designed for a single label structure. Therefore, we adjust the above three measures to fit our method. 

$\mathbf{F_{Ha}}$
 The traditional precision $P$ and recall $R$ rate are extended to the hierarchical precision $P_{H}$ and recall $R_{H}$ rate, which can well measure the severity of mistakes, as the error in the superclasses is more serious than that in the subclasses. As our MMF deals with multiple label structures, we take the average of all $P_{H}$ and $R_{H}$ in each label structure. $F_{Ha}$ is calculated from $P_{Ha}$ and $R_{Ha}$:
\begin{center}
\vspace{-12px}
\begin{equation} \nonumber
     P_{Ha}=\frac{1}{M} \sum_{m=1}^{M}\frac{\left | \hat{C_{aug}^{m}}\cap C_{aug}^{m} \right |}{\left | \hat{C_{aug}^{m}} \right |},\  R_{Ha}= \frac{1}{M} \sum_{m=1}^{M}\frac{\left | \hat{C_{aug}^{m}}\cap C_{aug}^{m} \right |}{\left | C_{aug}^{m} \right |} ,
\end{equation} 
\begin{equation}
 F_{Ha}= \frac{2\cdot P_{Ha}\cdot R_{Ha}}{P_{Ha}+R_{Ha}},
 \end{equation}
\end{center}
where $M$ is the number of the label structures,  $\hat{C_{aug}^{m}}$ is the predicted extension set which contains the class nodes on the path from the root class to the predict subclass in ${\mathbf{H_{m}}}$, $C_{aug}^{m}$ is the real extension set which contains the class nodes on the path from the root class to the real subclass in ${\mathbf{H_{m}}}$, and $|\cdot |$ is an operator to calculate the number of the elements.

$\mathbf{TIE_{a}}$
 In the tree structure, the total number of edges from the predicted node to the real node along a specific label structure is represented as $TIE$ distance. To deal with multiple label structures, we introduce ${TIE_{a}}$ to average all the $TIE$ distances in each label structure by Eq.(5), where $\left | Edge_{m}(c,\hat{c})\right |$ is the number of edges from the predicted node $\hat{c}$ to the real node $c$ in ${\mathbf{H_{m}}}$. Accordingly, the smaller the ${TIE_{a}}$, the more similar the predicted class is to the real class.
\vspace{-3px}
\begin{equation}
 TIE_{a} = \frac{1}{M}\sum_{m=1}^{M}\left | Edge_{m}(c,\hat{c})\right |.
\end{equation}

$\mathbf{LCA_{a}}$ We modified the $LCA$ height to the mean value of all $LCA$ heights in each label structure to obtain $LCA_{a}$ by Eq.(6), where $Height_{m}(c,\hat{c})$ is the lowest common ancestor height between the predicted node $\hat{c}$ and the real node $c$ in ${\mathbf{H_{m}}}$. A smaller $LCA_{a}$ means a smaller classification error. 
\vspace{-1px}
\begin{equation} 
LCA_{a}=\frac{1}{M}\sum_{m=1}^{M} Height_{m}(c,\hat{c}).
\end{equation} 
\vspace{-18px}

\subsection{Traing and Inference}
The multi-task loss for MMF model contains a CCB loss and several MSCBs losses denoted by Eq.(7), where $\mathit{\phi }(x;\theta)$ is a classification network, the parameter $\theta$ is learned by minimizing our loss function. $\hat{c}$ , $\hat{c_{}^{s_{m}}}$ are the predicted subclass and superclass, $c$, $c_{}^{s_{m}}$ are the ground truth of subclass and superclass in $\mathbf{H_{m}}$ respectively. $\lambda _{m}$ is the constraint intensity of ``$\mathbf{H_{m}}$ classifier", and $\lambda= \sum_{m=1}^{M}\lambda_{m}$. We use the standard cross entropy loss to compute $L_{CCB}$ and $L_{H_{m}}$.
 \vspace{-5px}
 \begin{equation}
 \mathcal{L}(\mathit{\phi } (x,\theta ),c,\mathbf{C_{}^{S}}) 
 = (1-\lambda )*L_{CCB}(\hat{c},c) \\ + \sum_{m=1}^{M} \lambda_{m}* L_{H_{m}}(\hat{c_{}^{s_{m}}},c_{}^{s_{m}}).
 \end{equation}
 \vspace{-5px}

When training, samples with different hierarchical label structures are input into the framework for multiple rounds of iterative training. MSCBs impose constraints on the network through the multi-task loss, affecting the prediction of subclasses.  When it comes to inference, the final predicted result of subclass is decided by CCB only.

\section{ Experiments}
\vspace{-3px}
\subsection{Experimental Settings}
{\bf Datasets}
We conduct experiments on two benchmark datasets CIFAR100 and Car196. In CIFAR100, there is a total number of 100 categories belonging to 20 semantic superclasses on average. Car196 is a fine-grained dataset containing 196 subclasses from three different kinds of semantic superclasses ``Make" (49 categories), ``Type"(18 categories), and ``Year". We choose ``Make" and ``Type" as the semantic label structures because ``Year" is not discriminative. We also construct a three-level $\mathbf{H_{A}}$ for each dataset. 

{\bf Backbones} The backbones of our network are VGG16 \cite{DBLP:journals/corr/SimonyanZ14a} and ResNet50 \cite{DBLP:conf/uemcom/VermaQF17} trained from scratch.  Note that there are five stages in both backbones. 
 
 {\bf Evaluation Metrics}
Four evaluation metrics are considered in our work to fully analyze the classifiers’ results. Besides the flat measure top-1 accuracy (Acc), we also adopt three hierarchical measures proposed before to better evaluate the performance of the classifiers.

\begin{table*}[htb]
\centering
\caption{The subclass performance with different $\rm \mathbf{H_{A}}$.}
\label{tab1}
\scalebox{0.9}{
\begin{tabular}{|c|c|cccc||cccc|}

\hline

\multirow{2}[2]{*}{Dataset}& \multirow{2}[2]{*}{$Num_{super}$} & \multicolumn{4}{c||}{VGG16} & \multicolumn{4}{c|}{ResNet50} \\\cline{3-10}
        & & Acc$(\uparrow)$& $F_{Ha}(\uparrow)$ & $TIE_{a}(\downarrow)$  & $LCA_{a}(\downarrow)$  &Acc$(\uparrow)$ & $F_{Ha}(\uparrow)$ &$TIE_{a}(\downarrow)$  &$LCA_{a}(\downarrow)$  \\   \hline
       
         \multirow{4}[2]{*}{CIFAR100}& 18&72.67&\textbf{84.15}&\textbf{0.9509}&\textbf{0.4754}&79.20&\textbf{88.12}&\textbf{0.7130}&\textbf{0.3565}\\
       &20&72.51&84.07&0.9558&0.4779&79.01&88.04	&0.7176	&0.3588\\
       &25&72.46&83.78&0.9733&0.4867&79.13&87.93	&0.7240	&0.3620\\
       &30&\textbf{72.95}&84.04&0.9574&0.4787&\textbf{79.21}&87.91&0.7252&0.3626\\

\hline

       \multirow{6}[2]{*}{Car196} & 15&\textbf{82.67}&\textbf{91.70}&\textbf{0.4979}&\textbf{0.2490}&\textbf{90.19}&\textbf{95.44}&\textbf{0.2738}&\textbf{0.1369}\\  
       &18&81.04&90.87&0.5478&0.2739&89.17&95.06&0.2963&0.1482\\
       &20&82.42&91.47&0.5116&0.2558&89.69&95.22&0.2865&0.1433\\
       &30&79.58&89.38&0.6370&0.3185&89.06&94.47&0.3321&0.1660\\
        &40&79.09&88.66&0.6804&0.3402&89.26&94.33&0.3402&0.1701\\
        &50&79.76&89.05&0.6569&0.3285&89.67&94.50&0.3297&0.1649\\
                  
\hline 
\end{tabular}
}

\end{table*}

\vspace{-5px}
\subsection{Ablation Study}
The impact of $\rm \mathbf{H_{A}}$, the network stages to attach MSCBs, and the constraint intensity $\lambda$ on the model's performance is explored in the following ablation experiments. Note that all the ablation studies are adapted both VGG16 and ResNet50 backbone on CIFAR100 and Car196, in order to show the generalization of our model.
\vspace{-15px}
\subsubsection{$\rm \mathbf{H_{A}s}$}  
As $\rm \mathbf{H_{A}}$ is three-level, the number of superclasses decides its structure. To obtain a suitable number of superclasses, we perform a series of ablation experiments on our MMF model with a singe label structure $\rm \mathbf{H_{A}}$. Referring to the number of superclasses in $\rm \mathbf{H_{S}}$, We vary the number of $\rm \mathbf{H_{A}}$'s superclasses in [18, 20, 25, 30] for CIFAR100 , and [15, 18, 20, 30, 40, 50] for Car196. According to the results of Table 1, we select the $\rm \mathbf{H_{A}}$ with 30 superclasses for CIFAR100, and 15 for Car196.
 \vspace{-5px}

 \begin{figure}[htb] 
 \center{\includegraphics[width=\textwidth] {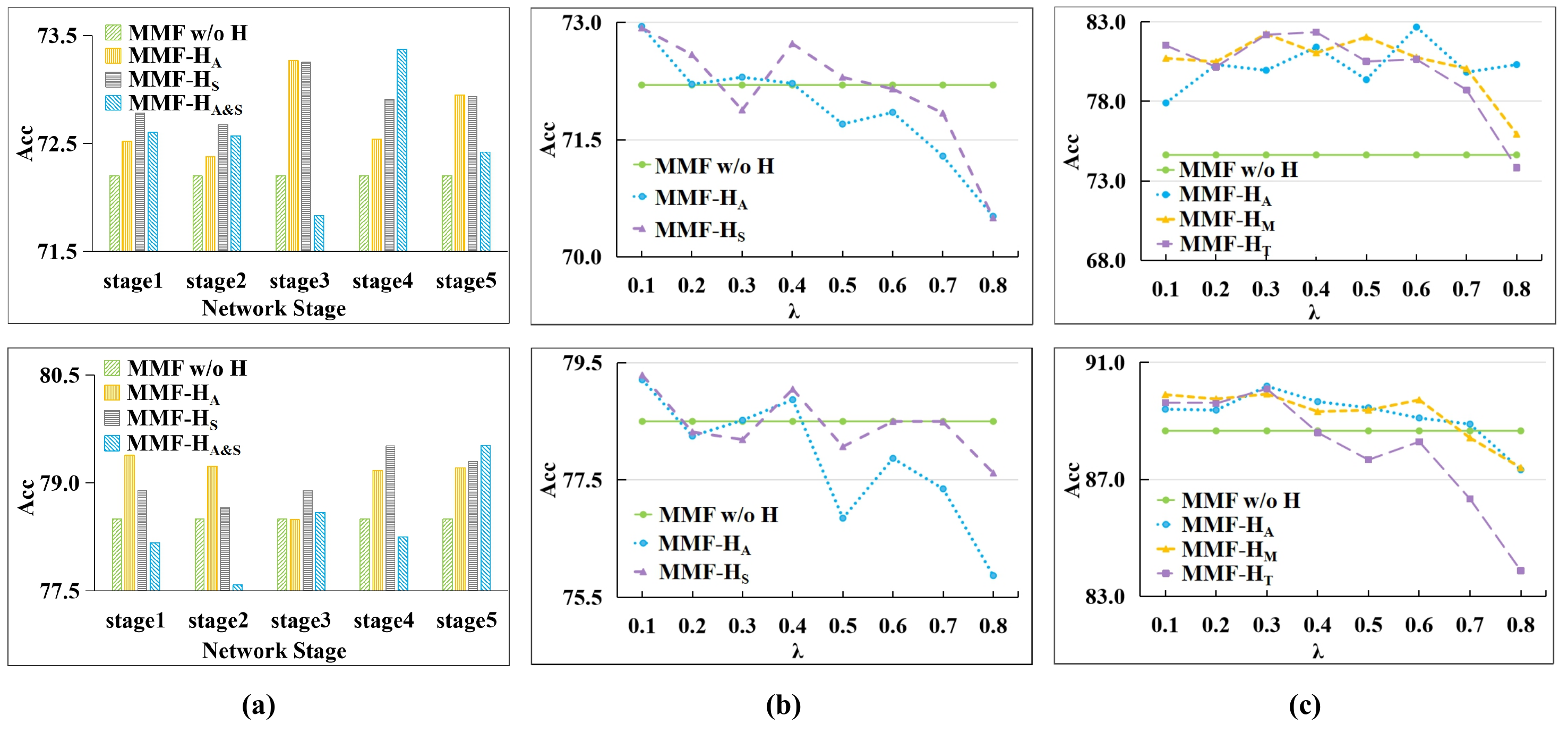}}
 \caption{\label{1} (a) MSCBs attached in different stages on CIFAR100. (b) and (c) are the constraint intensities of MSCBs on CIFAR100 and Car196 respectively. Note that results in the first row are for VGG16, and the second are for ResNet50.  }  
 \end{figure}

\begin{table*}[htbp]
  \centering
  \caption{The subclass performance of VGG16.}\scalebox{0.9}{
    \begin{tabular}{|c|c|c|c|c|cccc|}
    \hline
    Dataset & Method & Structure & $Num_{class}$ & $\lambda$&  Acc$(\uparrow)$ & $F_{Ha}(\uparrow)$ & $TIE_{a}(\downarrow)$ & $LCA_{a}(\downarrow)$ \\
    \hline
   \multirow{6}[2]{*}{CIFAR100} & Greedy \cite{DBLP:conf/nips/BengioWG10,DBLP:conf/cvpr/LiLC14} & $\rm \mathbf{H_{S}}$    & 100/20 &   -   &  70.09    &   81.98   &  1.0811    &  0.5405     \\
    \cline{2-9}
    & NBPath \cite{DBLP:journals/tip/QuLSLWXT17}& $\rm \mathbf{H_{S}}$    & 100/20 &  -    &  70.49    &   82.23   &   1.0664   &   0.5332   \\
    \cline{2-9}
    & \multirow{4}[2]{*}{MMF} & w/o $\rm \mathbf{H}$&100&-&72.20&83.35&0.9991&0.4995\\
    &   & $\rm \mathbf{H_{A}}$&100/30&0.1 &73.27&\textbf{84.97}&\textbf{0.9015}&\textbf{0.4507} \\
    &    &$\rm \mathbf{H_{S}}$ &100/20 &0.1 &73.25&84.70&0.9179&0.4590 \\
    &    & $\rm \mathbf{H_{A\&S}}$&100/30/20 &0.15 &\textbf{73.37}&84.79&0.9127&0.4563 \\
    \hline
    \hline
    \
    \multirow{12}[2]{*}{Car196} & \multirow{2}[2]{*}{Greedy \cite{DBLP:conf/nips/BengioWG10,DBLP:conf/cvpr/LiLC14}} & $\rm \mathbf{H_{T}}$ & 196/18   & -    & 51.45  & 73.89 & 1.5663 & 0.7832\\
    & & $\rm \mathbf{H_{M}}$    & 196/49 &  -  & 54.05 & 75.40 &1.4763  & 0.7381  \\
    \cline{2-9}
     &\multirow{2}[2]{*}{ NBPath \cite{DBLP:journals/tip/QuLSLWXT17}} & $\rm \mathbf{H_{T}}$ & 196/18   &    -  &  52.99 & 74.71 & 1.5172 & 0.7586 \\
    && $\rm \mathbf{H_{M}}$    & 196/49 &    -  &  55.30 & 76.05 & 1.4373 & 0.7186 \\
    \cline{2-9}
  
    &\multirow{8}[2]{*}{MMF} &  w/o $\rm \mathbf{H}$&196&-&74.63&87.88&0.7273&0.3637 \\ 
    &   &$\rm \mathbf{H_{A}}$&196/15&0.6&82.67&91.70&0.4979&0.2490 \\
    &   &$\rm \mathbf{H_{T}}$&196/18&0.4&82.35&91.34&0.5344&0.2672 \\
    &    &$\rm \mathbf{H_{M}}$&196/49&0.3&82.24&91.89&0.4864&0.2432 \\
    &    &$\rm \mathbf{H_{A\&T}}$&196/15/18&0.3&82.67&92.23&0.4661&0.2330\\
    &   &$\rm \mathbf{H_{A\&M}}$&196/15/49&0.3 &81.62&91.77&0.4938&0.2469 \\
    &    &$\rm \mathbf{H_{T\&M}}$&196/18/49&0.2
       &80.69&91.44&0.5134&0.2567 \\
    &   & $\rm \mathbf{H_{A\&T\&M}}$&196/15/18/49&0.2 &\textbf{83.67}&\textbf{92.88}&\textbf{0.4274}&\textbf{0.2137} \\
    \hline
    \end{tabular}}%
  \label{tab:addlabel}%
\end{table*}%
 
 \vspace{-25px}
\subsubsection{$\rm \mathbf{MSCB_{s}}$}
An important thing for our MMF model is where to insert the classifiers for superclasses. We explore it with $\rm \mathbf{H_{A}}$, $\rm \mathbf{H_{S}}$ and multiple structures $\rm \mathbf{H_{A\&S}}$ on CIFAR100, and the results with $\lambda = 0.2$ are shown in Fig.4(a). One interesting phenomenon can be observed in the both backbones: adding the superclass classifiers in the early stages is more effective. The reason may be that the low-level features are more generic and lose details of the high-level features for subclasses identification. So in the experiments, we insert MSCBs in the early stages to make our MMF model firstly grasp general concepts, then the CCB captures details in each concept to discriminate subclasses.

{\bf ${\lambda}$:} ~We fix MSCBs on the stage where the best performance is achieved, then vary $\lambda$ in [0.1, 0.8]. In Fig.4(b) and (c), experimental results on the subclasses show that different Acc obtained by adjusting $\lambda$. With a larger $\lambda$ ($\lambda \ge 0.1$), the performance on the subclasses is worse than the MMF w/o \textbf{H}. And it's not weird that results of different label structures don't coincide exactly because they have different similarity constraints, corresponding to different constraint strengths. In the following experiments with a single label structure, we set $\lambda$ with the best performance. And for multiple label structures, $\lambda_{m}$ is set to the same values for the sake of making these label structures act equally, varying within the range of the $\lambda$ which achieved the best results in the single label structures.

\begin{table*}[htbp]
  \centering
  \caption{The subclass performance of ResNet50.}\scalebox{0.9}{
    \begin{tabular}{|c|c|c|c|c|cccc|}
    \hline
   Dataset& Method & Structure & $Num_{class}$ & $\lambda$&  Acc$(\uparrow)$ & $F_{Ha}(\uparrow)$ & $TIE_{a}(\downarrow)$ & $LCA_{a}(\downarrow)$  \\
    \hline
   \multirow{6}[2]{*}{CIFAR100}& Greedy \cite{DBLP:conf/nips/BengioWG10,DBLP:conf/cvpr/LiLC14}& $\rm \mathbf{H_{S}}$    & 100/20 &    -   &   76.22   &  85.68    &   0.8594   & 0.4297 \\
    \cline{2-9}
   & NBPath \cite{DBLP:journals/tip/QuLSLWXT17}& $\rm \mathbf{H_{S}}$    & 100/20 &   -    &  76.44    &  85.78    &   0.8535   & 0.4267 \\
    \cline{2-9}
    &\multirow{4}[2]{*}{MMF} & w/o $\rm \mathbf{H}$&100&-&78.50&87.22&0.7668&0.3834\\
    &   & $\rm \mathbf{H_{A}}$&100/30 &0.1&79.38&88.14&0.7114&0.3557\\
    &    &$\rm \mathbf{H_{S}}$ &100/20 &0.1&79.51&88.28&0.7034&0.3517\\
    &    & $\rm \mathbf{H_{A\&S}}$&100/30/20 &0.15 &\textbf{79.52}&\textbf{88.28}&\textbf{0.7029}&\textbf{0.3514}\\
    \hline
    \hline
    \multirow{12}[2]{*}{Car196} & \multirow{2}[2]{*}{Greedy \cite{DBLP:conf/nips/BengioWG10,DBLP:conf/cvpr/LiLC14}} & $\rm \mathbf{H_{T}}$ & 196/18    &  -   & 86.80  & 93.30 & 0.4022 & 0.2011 \\
   & & $\rm \mathbf{H_{M}}$    & 196/49 & - & 87.30 & 93.57 & 0.3855 & 0.1928 \\
    \cline{2-9}
    & \multirow{2}[2]{*}{NBPath \cite{DBLP:journals/tip/QuLSLWXT17}} & $\rm \mathbf{H_{T}}$ & 196/18   &  -    & 87.40  & 93.63 & 0.3823 & 0.1911 \\
    && $\rm \mathbf{H_{M}}$    & 196/49 & - & 87.69  & 93.78 & 0.3732 & 0.1866 \\
    \cline{2-9}
  
    &\multirow{8}[2]{*}{MMF} &  w/o $\rm \mathbf{H}$&196&-&88.66&94.84&0.3097&0.1548 \\ 
    &   &$\rm \mathbf{H_{A}}$&196/15&0.3&90.19&95.44&0.2738&0.1369\\
    &   &$\rm \mathbf{H_{T}}$&196/18&0.3&90.10&95.44&0.2736&0.1368\\
    &    &$\rm \mathbf{H_{M}}$&196/49&0.3&89.92&95.59&0.2645&0.1322\\
    &    &$\rm \mathbf{H_{A\&T}}$&196/15/18&0.1&90.20&95.72&0.2567&0.1283\\
    &   &$\rm \mathbf{H_{A\&M}}$&196/15/49&0.1&89.45&95.41&0.2756&0.1378\\
    &    &$\rm \mathbf{H_{T\&M}}$&196/18/49&0.2/0.1 &\textbf{90.42}&\textbf{95.89}&\textbf{0.2468}&\textbf{0.1234}\\
    &   & $\rm \mathbf{H_{A\&T\&M}}$&196/15/18/49&0.05 &90.29&95.87&0.2477&0.1239\\
    \hline
    \end{tabular}}%
  \label{tab:addlabel}%
  
\end{table*}%

\subsection{Experimental Results and Analyses}

Our deep MMF model with different single label structures and their combinations is compared with two methods based on the top-down strategy. For the top-down methods, we choose two methods which are based on the greedy selection at each hierarchy (Greedy) \cite{DBLP:conf/nips/BengioWG10,DBLP:conf/cvpr/LiLC14} and the N-Best Path (NBPath) \cite{DBLP:journals/tip/QuLSLWXT17}. To improve the performance, we adopt features extracted from a carefully fine-tuned VGG16 or ResNet50, which is the backbone in our MMF model. Then kernel SVMs are employed as the classifiers at each hierarchy. For our MMF model, we adopt different label structures as shown in Table 2 and Tabel 3.
``w/o $\rm \mathbf{H}$" means MMF model without any hierarchical structures, which is a traditional classification network contains a backbone and a classifier for subclass classification. ``$\rm \mathbf{H_{T}}$" and ``$\rm \mathbf{H_{M}}$" are the semantic structures based on ``Type" and ``Make" respectively, and ``$\rm \mathbf{H_{A\&S}}$" (i.e., $\rm \mathbf{H_{A}}$ and $\rm \mathbf{H_{S}}$) $etc.$ are multiple label structures.

 Table 2 and Table 3 show the results of VGG16 and ResNet50 on CIFAR100 and Car196, respectively. Note that in the multi-structure models, the performance of the subclass classifiers achieves the best performance when the $\lambda_{m}$ for different structures is equal, except for $\rm \mathbf{H_{T\&M}}$ in ResNet50. It can be concluded that: 1) For the subclass classifier performance, our MMF model with a single structure is better than the top-down methods with a considerable margin, which verifies the efficiency of the end-to-end training. 2) Besides, MMF with any single structure achieves better performance than ``w/o $\rm \mathbf{H}$", indicating the benefit of the superclass classifiers.  3) Furthermore, MMF with multiple label structures performs better than any single one, which confirms our assumption that multiple label structures can provide richer similarity constraints to improve the performance of the subclass classifier.  4) The gain in hierarchical evaluation metrics is more obvious than the flat measure Acc, indicating that predictions in our MMF model are more closer to the ground truth (i.e., a less serious mistake).
\vspace{-10px}

\section{Conclusion}
\vspace{-5px}
In this paper, we have constructed a multi-task multi-structure fusion model for hierarchical classification. Various factors have been explored, such as different label structures based on the affinity matrix, the stages to attach the superclass classifiers, and theconstraint intensities. Besides, the hierarchical evaluation metrics have been adjusted to fit the classification with multiple label structures. The experimental results demonstrate that different label structures provide various prior knowledge for the subclass classifier. Meanwhile, integrating these multiple label structures can achieve better results.

In this work, relations of the subclass and its superclasses are impplicitly modeled by the weighted multi-task loss function. In the future, we will explore more direct ways to utilize multiple label structures. 

%

\bibliographystyle{splncs04}
\bibliography{icann}

\end{document}